# StairNet: Visual Recognition of Stairs for Human-Robot Locomotion

Andrew Garrett Kurbis,[1-3] Dmytro Kuzmenko,[4] Bogdan Ivanyuk-Skulskiy,[4] Alex Mihailidis,[1-3] and Brokoslaw Laschowski[2-3,5]

[1]Institute of Biomedical Engineering, University of Toronto, Toronto, Canada
[2]Robotics Institute, University of Toronto, Toronto, Canada
[3]KITE Research Institute, Toronto Rehabilitation Institute, Toronto, Canada
[4]Department of Mathematics, National University of Kyiv-Mohyla Academy, Kyiv, Ukraine
[5]Department of Mechanical and Industrial Engineering, University of Toronto, Toronto, Canada

## Abstract

Human-robot walking with prosthetic legs and exoskeletons, especially over complex terrains such as stairs, remains a significant challenge. Egocentric vision has the unique potential to detect the walking environment prior to physical interactions, which can improve transitions to and from stairs. This motivated us to create the StairNet initiative to support the development of new deep learning models for visual sensing and recognition of stairs, with an emphasis on lightweight and efficient neural networks for onboard real-time inference. In this study, we present an overview of the development of our large-scale dataset with over 515,000 manually labeled images, as well as our development of different deep learning models (e.g., 2D and 3D CNN, hybrid CNN and LSTM, and ViT networks) and training methods (e.g., supervised learning with temporal data and semi-supervised learning with unlabeled images) using our new dataset. We consistently achieved high classification accuracy (i.e., up to 98.8%) with different designs, offering trade-offs between model accuracy and size. When deployed on mobile devices with GPU and NPU accelerators, our deep learning models achieved inference speeds up to 2.8 ms. We also deployed our models on custom-designed CPU-powered smart glasses. However, limitations in the embedded hardware yielded slower inference speeds of 1.5 seconds, presenting a trade-off between human-centered design and performance. Overall, we showed that StairNet can be an effective platform to develop and study new visual perception systems for human-robot locomotion with applications in exoskeleton and prosthetic leg control.

1. Introduction

Robotic leg prostheses and exoskeletons can provide locomotor assistance to individuals affected by impairments due to aging and/or physical disabilities such as stroke [1]. Most control systems for human-robot walking use a hierarchical strategy with high, mid [2], and low [3] level controls. Robotic leg control requires continuous assessment of locomotor states for seamless transitions between different operating modes. Previous high-level controllers relied on mechanical, inertial, and/or electromyographic (EMG) sensors for state estimation, which are generally limited to the current state, analogous to walking blind. Inspired by the human vision system [4], [5], egocentric vision can uniquely detect environmental states prior to physical interaction and thus aid in smooth and accurate transitions. However, the classification of walking terrains such as stairs presents additional challenges because of the complex nature of real-world environments, which can vary significantly in style, material, and geometry. The classification of stairs is particularly important because of the increased risk of severe injury from falls if the environment is misclassified.

Previous vision systems have been developed to recognize stairs for robotic leg control using hand-designed feature extractors [6]–[10] or automated feature engineering via convolutional neural networks (CNNs) [11]–[19]. However, these systems have inherent limitations in terms of performance and generalizability to new environments because of suboptimal hand engineering and/or training on relatively small image datasets. Recent studies have significantly expanded the number of labeled images [20] and presented the opportunity to use deep learning models to increase performance and generalizability.

The purpose of this study is to provide an overview of our StairNet initiative, which we created to support the development of new deep learning models for visual sensing and perception of stair environments for human-robot walking. The initiative emphasizes lightweight and efficient neural networks for onboard real-time deployment on mobile and embedded devices. We discuss the development of our large-scale dataset with over 515,000 manually labeled images, as well as our development of different deep learning models and training methods using our new dataset. Building on this work, the StairNet initiative can support the development of next-generation environment-adaptive control systems for robotic leg prostheses, exoskeletons, and other assistive technologies for human locomotion.

2. StairNet Dataset

Our StairNet dataset contains over 515,000 RGB images that we manually annotated using hierarchical class labels for environments encountered during level-ground and stair locomotion. To our knowledge, this dataset is the largest and most diverse dataset of egocentric images of stair environments published to date. We made the dataset open source at https://ieee-dataport.org/documents/stairnet-computer-vision-dataset-stair-recognition to support the research community and allow for direct comparisons between different machine learning models. The dataset includes annotated class labels to reduce class overlap and maximize the theoretical performance for model development.

We developed the StairNet dataset using images from ExoNet [20], which were captured using a chest-mounted wearable camera (iPhone XS Max) in indoor and outdoor environments. The images were saved at 5 frames/s with a resolution of 1280x720 pixels with multiple users with varying heights and camera pitch angles. In our initial study, we found that the ExoNet labels contained many overlapping classes, which resulted in limited performance for models trained using these annotations [12]. Therefore, we developed new class definitions to manually re-label the images and to increase the precision of the cut-off points used between the different walking environments. We defined four new classes, including level-ground (LG), level-ground transition to incline stairs (LG-IS), incline stairs (IS), and inclined stairs transition to level-ground (IS-LG). We performed three manual labeling pass-throughs to increase annotation accuracy and precision. We removed images that did not contain either level-ground terrain or incline stairs or had significant camera obstructions. Since our dataset is designed for stair recognition, there is no loss of characteristics related to the intended application by removing these images, as any classifications made outside of these classes are considered out of scope and would require additional models for classification.

Our dataset repository also includes information related to the class distribution and definitions. The dataset mainly comprises images of level-ground terrain (86% of samples) and incline stairs (9%), with two minority classes, IS-LG and LG-IS, which contain approximately 2% and 3% of the samples, respectively. This imbalance is important to consider when selecting classification and resampling methods. When using our dataset for model development, we suggest using a video-based train-validation-test split, as outlined in [20]. This method assigns all frames within a video episode (i.e., group of neighboring frames) to one of the dataset splits to prevent data leakage and to provide a better estimation of real-world performance and generalizability [21]. Scripts for this validation approach and data preprocessing can be found on our GitHub.

Using the StairNet dataset, we developed and tested a number of different deep learning models and training methods to directly evaluate and compare their advantages and disadvantages on a common platform, as subsequently discussed.

## 3. Deep Learning Models

### 3.1 Baseline Model

The first StairNet model [12], also known as our baseline model, was developed using supervised learning, which predicted each frame independently, as shown in Figure 1. We developed an efficient 2D CNN based on the architecture of MobileNetV2 for image classification, which was designed by Google for mobile and embedded vision applications [22], [23]. MobileNetV2 uses depth-wise separable convolutions with width and resolution multipliers to create a lightweight framework with the trade-off of slightly lower accuracy for significant reductions in computational requirements, which is suitable for onboard real-time inference for robot control.

We developed the baseline model using TensorFlow 2.7 [24], starting with the default parameter values from [18], [25], [26]. We used a Google Cloud Tensor Processing Unit (TPU) to help efficiently optimize model parameters. A global average pooling 2D layer and softmax dense prediction layer were added for transfer learning with pretrained weights from ImageNet [27]. Five freeze layer hyperparameters were tested: 141, 100, 50, 25, and 5, with each variation trained for 60 epochs. Five frozen layers with 2.2 million trainable parameters resulted in the highest validation accuracy and lowest validation loss. A grid search found an optimal combination of a batch size of 256 and a learning rate of 0.0001. Using these hyperparameters, pretrained weights were compared with randomly initialized weights. After 60 epochs, both validation accuracy curves plateaued, with the pretrained model outperforming the randomly initialized model with validation accuracies of 98% and 97%, respectively. However, characteristics of overfitting were observed for both models. To address this, additional regularization was implemented via a dropout layer with dropout rates between 0.1 and 0.5, and L2 weight regularization. A dropout rate of 0.2 resulted in reduced overfitting and an increase in validation performance, while additional L2 weight regularization was removed with no impact on model performance.

To further reduce overfitting, we oversampled the underrepresented transition classes (IS-LG and LG-IS). Images were randomly resampled and augmented during training with five oversampling values (i.e., 25,000, 40,000, 60,000, 200,000, and 400,000) to control the minimum number of images per class. Our experiment showed that a higher minimum value per class decreased the overall validation accuracy. However, the categorical accuracy for the underrepresented classes increased, creating a more even categorical accuracy distribution across the different walking environments. Given that more significant consequences could result from a false negative than a false positive for human-robot locomotion, a minimum value of 400,000 images per class was used to minimize the probability of false negatives, as seen in the increased accuracy in the IS and IS-LG classes, with increases of 0.3% and 2.2%, respectively.

The baseline model underwent a final round of hyperparameter optimization for batch size and learning rate in a high epoch run. After multiple iterations, we finalized the model using a reduced base learning rate of 0.00001, a batch size of 128, and a cosine weight decay learning policy. The model included pretrained weights, five frozen layers, 2.3 million parameters, and a minimum categorical image count of 400,000 images. We also added a dropout layer with a dropout rate of 0.2. The final model was trained for 100 epochs with early stopping.

The model was evaluated using the train, validation, and test sets of the StairNet dataset described in Section 2. The model achieved 99.3% and 98.5% accuracies on the training and validation sets, respectively. When evaluated on the test set, the model achieved an overall classification accuracy of 98.4%, correctly classifying 35,507 of the 36,085 images. Additionally, the model achieved an F1 score of 98.4%, weighted precision value of 98.5%, and weighted recall value of 98.4%. The model achieved this performance with 2.3 million parameters and 6.1 GFLOPs. The classification accuracy on the test set varied between different environments, with categorical accuracies of 99.0% for LG, 91.7% for LG-IS, 96.9% for IS, and 90.5% for IS-LG. The two transition classes (i.e., LG-IS and IS-LG), comprising only 3.1% and 1.8% of the total number of images, respectively, achieved the lowest categorical accuracies.

Our baseline model had failure cases such as incorrectly predicting a transition to incline stairs when level-ground images contained strong horizontal lines in the top section of the image. Images with strong horizontal lines throughout the image (e.g., brick flooring or tiles) also presented difficulties for the model and led to incorrect classification of incline stairs. False negatives were less common but occurred from encountering unique stair characteristics such as unusual materials (e.g., a stair under repair with a wood plank over the base material) or viewing angle (e.g., looking to the left or right while walking upstairs). We used this baseline model as a reference and benchmark for our subsequent models that we developed and studied.

3.2 Mobile Deployment

To evaluate the real-world performance of our baseline model, we custom-designed a mobile app using TensorFlow Lite (TFLite) [28] and Swift 5 and Xcode 13.4.1 [29] for on-device inference [13]. The mobile app prepares an image from the camera feed and scales the input resolution using a square crop to match the resolution of our deep learning model input size (i.e., 224x224). The model then runs on-device inference, outputting the tensor results in a float-array format containing the confidence values for the four walking environments for each image. The mobile interface displays the output information with the class predictions, along with the onboard inference speed (ms) for the last image.

We use a TFLite interpreter for the on-device computation, which has several advantages over other deployment methods such as cloud computing. It allows offline execution and inference on edge devices without requiring an internet connection or the need to communicate with a machine learning server. Performing offline inference can significantly reduce power requirements and privacy concerns, particularly in clinical applications, as no data is required to leave the device. TFLite also has a small binary size and supports highly efficient models for low inference times, with minimal impact on accuracy during compression.

For mobile deployment, the baseline model was converted from its original h5 format to a TFLite flat buffer format. This conversion allows for onboard processing and inference via the on-device interpreter and built-in TFLite infrastructure (see Figure 2), which supports multiple backend processing options such as central processing units (CPUs), graphics processing units (GPUs), and neural processing units (NPUs). We experimented with five different conversion methods with varying degrees of compression, which can increase inference speed at the expense of accuracy. These compression formats included: 1) Float32 compression, the default format for general TFLite deployment; 2) post-training float16 quantization, which reduces the model size and boosts its performance on hardware with optimized float16 computation; 3) post-training int8 weight quantization, which reduces the model size and improves performance on CPU hardware; 4) post-training quantization with int16 activations to reduce the model size and make it compatible with integer-only accelerators; and 5) post-training int8 full model quantization (i.e., model weights, biases, and activations), which reduces the model size and increases processor compatibility. Each compression format was evaluated using the StairNet test set to determine the effect of model compression on accuracy.

We tested the inference speeds of our baseline model on four different mobile devices (i.e., iPhone 8+, iPhone X, iPhone 11, and iPhone 13) with four different backend processing options, including a single-threaded CPU, a multithreaded CPU, GPU, and a combination of CPU, GPU, and NPU. We developed these backend processing options using APIs with access to hardware accelerators. These APIs included the Apple Metal delegate for direct GPU compute and the Apple CoreML delegate, which uses the three iOS processing options to maximize performance while minimizing memory usage and power consumption. An offline test was performed on each device and backend processing option using a pre-recorded video, eliminating variation in camera input on the inference speed test. The pre-recorded video contained stair ascent in indoor and outdoor environments and was loaded to the mobile app to mimic the camera feed. The average inference time was calculated using inference times sampled at 5-second intervals during the video for each experiment.

When compressed for mobile deployment, our baseline model had accuracy reductions between 0.001-0.111% compared to the full-sized model. The compressed model formats of float32 and float16 quantization resulted in the highest accuracy post-conversion (98.4%). In contrast, the int8 quantization format with both int8 and int16 activations had the lowest post-conversion accuracies of 98.3% and 98.3%, respectively.

The model achieved an inference speed of 2.75 ms on our mobile app using the CoreML delegate and float32 model. The Core ML and Metal delegates, which use parallel processing of CPU, GPU, and NPU, and direct GPU compute, performed best on newer devices such as the iPhone 11 and iPhone 13. The inference times for these devices were 2.75 ms and 3.58 ms, respectively. In contrast, CPU processing resulted in slower inference times of 9.20 ms and 5.56 ms when using single and multithreaded CPUs. On older devices such as iPhone 8+ and iPhone X, multithreaded CPU achieved faster inference times when compared to single-threaded CPU and GPU processing. When using the CoreML delegate, the float32 compression format delivered the fastest inference speed across all devices. Similarly, the float32 format achieved the fastest inference speeds when running on a GPU with metal delegate. For mobile CPU performance, int8 quantization with int16 model activations resulted in the fastest inference time for single and multithreaded processing, with average speeds of up to 9.20 ms and 5.56 ms, respectively.

Accordingly, we developed a baseline model using the StairNet dataset and deployed the model on our custom-designed mobile app for stair recognition, achieving high classification accuracy and low latency. However, this research was limited to standard supervised learning and did not take into consideration the temporal nature of human-robot walking, which motivated our subsequent research.

3.3 Temporal Neural Networks

To study the effect of sequential inputs on classification performance compared to our baseline model, which used independent frames, we developed several temporal neural networks [30] to exploit information from neighboring frames in the StairNet dataset (see Figure 3). We experimented with different deep learning models, including the new lightweight 3D CNN architecture called

MoViNet [31], and a number of hybrid encoder architectures, including VGG-19 [32], EfficientNet-B0 [33], MobileNetV2 [23], MobileViT [34], and ViT-B16 [35], each paired with a temporal long-short term memory (LSTM) backbone [36], and a transformer encoder [37]. We performed focused testing on the 3D MoViNet model, MobileViT with LSTM, and MobileNetV2 with LSTM, which were selected based on their potential to accurately recognize images of stairs and capture temporal dynamics.

We first experimented with MoViNet- a modified version of MobileNetV3 designed for videos. We used MoViNet's neural architecture search (NAS) to optimize the model parameters such as the number of layers, convolutional filter width, and number of feature map channels. To reduce the growth of model memory, we implemented a stream buffer to act as a cache feature applied to the boundaries of the video subsequences. The cache was zero-initialized. To compute the feature map, we applied a temporal operation (i.e., 3D convolution) over the concatenation of the buffer and the subsequence. The buffer was updated in subsequent feature maps by concatenating the current buffer and new sequence using the following formula:

$$B_{i+1} = \left(B_i x_i^{clip}\right)_{[-b:]} \quad (1)$$

where $B_i$ is the buffer, $x_i^{clip}$ is the original input sequence, and $[-b:]$ is the selection of the last $b$ frames of the concatenated feature sequence. We used a stream buffer to reduce the memory use of the MoViNet model at the expense of a small reduction in accuracy. However, we mitigated this loss in accuracy by using an ensemble of models with two identical MoViNet architectures at a half-frame rate. During inference, the input sequence was fitted to both networks and the mean values of the two models were obtained and passed through the softmax activation function.

We also experimented with MobileNetV2 combined with LSTM. Similar to our baseline model, the MobileNetV2 architecture was chosen for its efficient model design, optimized for mobile and embedded devices. MobileNetV2 was applied to each frame of the sequence, resulting in a stack of feature maps, which was then fed into an LSTM layer to capture temporal dynamics. The output of the LSTM layer was a sequence of labels for sequence-to-sequence classification or the last predicted label of the LSTM recurrence operation for sequence-to-one classification.

Lastly, we experimented with MobileViT, a hybrid encoder model that combines local information from convolutional layers and global information using MobileViT blocks. The model has convolution layers represented by a stride of 3×3 convolution on the input and several MobileNetV2 blocks. The MobileViT blocks in the later layers extract feature maps that are used to encode global information. Standard convolutions are applied to encode the local spatial information, and point-wise convolutions are used to project to a high-dimensional space. These high-dimensional projections were then unfolded into non-overlapping flattened patches and encoded using transformer blocks. The transformer outputs were projected back to the original low-dimensional space and fused with the original feature maps.

Similar to MobileNetV2, the MobileViT model was applied to each frame of the sequence. This resulted in a sequence of feature maps, with each map corresponding to one frame. These feature maps were then passed through the transformer layer to capture temporal dynamics of the feature maps of each sequence. In sequence-to-sequence classification, the output of the last transformer block passed through a linear classification head. In sequence-to-one classification, we flattened the transformer layer output before the classification head.

We performed hyperparameter optimization using KerasTuner. The hyperparameter space for each group of models was selected based on the experimental setup and architecture. Once the optimal hyperparameters were determined, each model was trained for 20 epochs using an NVIDIA Tesla V100 32GB GPU. The Adam optimizer [38] was used with a learning rate of 0.0001, along with a cosine annealing learning rate scheduler. We used NetScore [39] to compare the optimized models, which balances the network performance with efficiency and is represented by the following equation:

$$\Omega(N) = 20 \log \frac{acc(N)^{\alpha}}{param(N)^{\beta} \, flops(N)^{\gamma}} \qquad (2)$$

where $acc(N)$ is the classification accuracy (%), $param(N)$ is the number of model parameters, which is indicative of the memory storage requirements, $flops(N)$ is the number of floating point operations, which is indicative of the computational requirements, and $\alpha, \beta, \gamma$ are coefficients that control the influence of each parameter on the overall NetScore.

We assessed the sequence-to-one models as regular classification models, in which the predicted label of the sequence was compared to the ground truth. Sequence-to-sequence models were evaluated in two ways. The first method was sequence-to-sequence evaluation, in which the predicted sequence of labels was compared to the ground truth of the sequence of labels. We also compared the sequence-to-one evaluated with the anchor frame label, as was used to compare the performance of sequence-to-sequence models with sequence-to-one models.

Of the temporal neural networks that we studied, the 3D MoViNet model achieved the highest classification performance on the StairNet test set, with 98.3% accuracy and an F1-score of 98.2%. The hybrid models, which contain a 2D-CNN encoder and temporal blocks (i.e., MobileNetV2 with LSTM and MobileViT with LSTM), struggled to capture inter-frame dependencies with minimal sequences (i.e., five frames per sample) [40] and thus achieved lower classification performance compared to our 3D model. The 3D model had the highest NetScore of 167.4, outperforming the 2D encoder models with scores of 155.0 and 132.1 for MobileViT with LSTM and MobileNetV2 with LSTM, respectively.

We calculated a NetScore of 186.8 for our baseline model, outperforming all temporal neural networks that we studied in terms of efficiency due to its relatively low number of parameters and numerical operations. Among the hybrid models, MobileViT with LSTM had slightly lower classification performance compared to MobileNetV2 with LSTM with F1-scores of 96.8% and 97.0%, respectively. However, the hybrid MobileViT model had a much higher NetScore, with disproportionally less parameters (3.4 million) and operations (9.8 billion FLOPS) compared to 6.1 million parameters and 54 billion FLOPS for hybrid MobileNetV2. We also showed an increase in performance using sequence-to-one methods on sequence-to-sequence models over the standard sequence-to-sequence method, with an accuracy of 97.3% and 70.7%, respectively, using the same sequence-to-sequence model.

In summary, of the temporal neural networks that we studied using sequential images for stair recognition, we showed that the 3D model outperformed the 2D models with temporal backbones in terms of classification accuracy and efficiency, which takes into consideration the computational and memory storage requirements. We also showed that the 3D video model achieved a higher classification accuracy (98.3%) compared to our 2D baseline model when retested on the video-based StairNet test set (97.2%). However, the 3D model had a lower NetScore (i.e., less efficient) due to disproportionally more parameters and operations.

3.4 Semi-Supervised Learning

Compared to the aforementioned research, all of which relied on standard supervised learning, we wanted to study the use of semi-supervised learning [41] to improve training efficiency by using large amounts of unlabeled data. Our acquisition and manual labeling of hundreds of thousands of images to develop the StairNet dataset in Section 2 was time-consuming, labour-intensive, and a significant bottleneck in the development of our initial deep learning models. Using large amounts of publicly available unlabeled data [20] is a viable option to increase training efficiency. The purpose of this work was to show the potential to improve efficiency by reducing the number of labeled images required for stair recognition while maintaining performance compared to our baseline model.

We used unlabeled images from ExoNet that were not included in the StairNet dataset. However, using unlabeled data can present challenges, including lack of information about the class distributions and viability of the images. We performed a visual search of the images and found that the unlabeled data had limitations similar to those in the StairNet dataset [12], [13], with images containing environments that were not relevant to stair recognition (i.e., outside of the four StairNet classes) and had significant camera obstructions. We used the FixMatch semi-supervised learning algorithm [42] due to its intuitive and feasible implementation compared to more complex algorithms such as self-training with noise students [43], meta-pseudo-labels [44], AdaMatch [45], and contrastive learning for visual representation [46].

Our semi-supervised pipeline consisted of three major steps (Figure 4): 1) labeled and unlabeled raw images were loaded and oversampled from the labeled dataset with augmentations to help mitigate false positives during training; 2) unlabeled image logits were retrieved using a supervised pretrained model to preprocess the unlabeled dataset. The most probable pseudo-labels were then selected if they surpassed the cutoff parameter $\tau$. Weak augmentations (i.e., horizontal flips) and strong augmentations (i.e., color intensity, saturation, small rotations, and horizontal flips) were applied to the images. The batch size ratio parameter $\mu$ is the difference between the labeled and unlabeled batch sizes. During training, the unlabeled data required a larger batch size than the labeled dataset. The labeled and unlabeled batches were used as inputs, inferred using weakly augmented images, and the received logits were then thresholded using a pseudo-label cut-off parameter; 3) the models were trained using a supervised loss (i.e., cross-entropy loss) and unsupervised loss (i.e., cross-entropy loss of the thresholded pseudo-label logits calculated against strong augmented images). The weight of the unsupervised loss on training was adjusted using the parameter $\lambda$. These semi-supervised parameters ($\tau$, $\lambda$, and $\mu$) were tuned to provide a high degree of model flexibility.

As previous research has shown that automated feature extractors are superior to handcrafted features, particularly on large-scale image datasets [21], we used convolutional and transformer-based architectures for our model development. We first developed a vision transformer model with the base architecture of MobileViT [34], which uses automated feature engineering similar to standard CNNs. MobileViT, which was also used in Section 3.3, is a transformer-based model that employs mechanisms of attention and depth-wise dilated convolution. The model uses low-level convolution and transformer blocks, allowing for high efficiency and inference speed similar to the lightweight CNN used in our baseline model [12], [13]. We tested three different backbones for MobileViT (i.e., XXS, XS, and S), which varied in terms of the number of transformer layers, more sophisticated feature extraction, and parameter count, allowing for an optimal trade-off between model size and performance. We developed our model using TensorFlow 2.0 and trained using a high-performance Google Cloud TPU.

Using the same StairNet dataset split distribution as our baseline model [12], [13], we reduced the labeled training data from 461,328 to 200,000 images to study the impact of reduced annotations. To address the issue of unknown class distribution and image quality of the unlabeled data, we used our supervised baseline model to retrieve the logits of the 4.5 million unlabeled images from ExoNet, which were thresholded using the FixMatch approach.

After processing the unlabeled dataset, 1.2 million images surpassed the 0.9 $\tau$ cut-off threshold. The resulting subset of images within the threshold had a pseudo-label distribution that closely resembled the original StairNet Dataset [12], [13] (i.e., 5.5% for IS, 1% for IS-LG, 90.1% for LG, and 3.4% for LG-IS). The lightest MobileViT XXS model was the fastest to train and infer among the three variants but had low accuracy during training. The balanced MobileViT XS backbone provided the best trade-off between model compactness and performance. The largest MobileViT S with 4.9 million parameters had the slowest training and inference times, while having worse overall performance likely due to overfitting.

Our best semi-supervised learning model was MobileViT XS, pretrained on ImageNet. The model was trained using stochastic gradient descent (SGD) with 0.9 momentum, a batch size of 64, Nesterov acceleration, randomly initialized, and with FixMatch parameters of $\tau$ = 0.98 and a loss weight of $\lambda$=1. During training, the data imbalance of both the labeled and unlabeled data was handled by replacing standard cross-entropy with a focal loss class weight penalization of $\gamma$ = 3 to penalize hard negatives. We also tested the exponential moving average, which averaged the parameters and produced significantly better results than the final weight matrices. The resulting model showed good convergence, but the overall image validation accuracy was inferior to that of the previous vanilla cross-entropy loss experiments.

To reduce the number of false positives, augmentations were implemented on the labeled training set, including minor translations, rotations, contrast, and saturation. Variations were tested in the L2 parameter loss and decoupled weight decay [47]. Our best models did not include weight decay regularization. We experimented with both cosine weight decay, as suggested by FixMatch [42], and cosine decay with restarts [48]. The former was found to be more resilient and consistent and thus was implemented in our final model. Several experiments were conducted to determine the optimal ratio of labeled to unlabeled data ($\mu$) and the unsupervised loss weight parameter ($\lambda$). The final model was trained using 300,000 labeled images (i.e., approximately 65% of the original training set) and 900,000 unlabeled images. The model had a MobileViT XS backbone and was optimized using SGD with Nesterov. The final set of hyperparameters was a learning rate of 0.045, pseudo-label cut-off $\tau$ of 0.9, a supervised batch size of 64, a batch size ratio $\mu$ of 3, an unsupervised loss weight $\lambda$ of 1.03, and a cosine decay learning rate schedule. To address class balancing, focal loss was replaced with categorical cross-entropy loss. The final model was trained for 42 epochs.

Our semi-supervised learning model achieved classification accuracies of 99.2% and 98.9% on the StairNet training and validation sets, respectively. When evaluated on the test set, the model achieved an overall image classification accuracy of 98.8%, a weighted F1-score of 98.9%, a weighted precision value of 98.9%, and a weighted recall value of 98.8%. Similar to our baseline model, the two transition classes (LG-IS and IS-LG) achieved the lowest categorical accuracies (90.6% and 90.4%), which can be attributed to having the smallest class sizes. Overall, our semi-supervised learning model achieved a similar image classification performance on the StairNet dataset as our baseline model [12], [13] but used 35% fewer labeled images, therein improving the training efficiency.

3.5 Embedded Deployment

Lastly, building of the aforementioned studies, we developed an integrated smart glasses solution to move towards a more human-centred design [49]. One of the limitations of our previous deep learning models was their use of images from a chest-mounted smartphone camera. These images do not necessarily coincide with the user's visual field, and thus are more difficult to infer intent, and are susceptible to obstructions such as the user's arms. However, previous head-mounted cameras [50]–[52] have mainly been limited to off-device inference using desktop computers and cloud computing. An integrated visual perception system has yet to be designed, prototyped, and evaluated on edge devices with low inference speeds. This gap could be explained by limitations in embedded computing, which have only recently been alleviated by advances in hardware and deep learning model compression methods.

Therefore, the purpose of this work was to develop a novel pair of AI-powered smart glasses that uniquely integrate both sensing and computation for visual perception of human-robot walking environments while achieving high accuracy and low latency. We integrated our mechatronic components all within a single device, which is lightweight and small form factor as to not obstruct mobility or user comfort. Computationally, it has sufficient memory and processing power for real-time inferencing with live video stream. We custom-designed 3D-printed mounts to allow our system to attach and be transferable to a wide range of eyeglass frames. The main mechatronic components is a lightweight camera to sense the walking environment and a microcontroller to process and compute the images. Inspired by commercial smart glasses such as Google Glass [50] and Ray-Ban Stores [51], our design features a forward-facing camera aligned with the user's field of view (i.e., egocentric), with the computational processing on the side of the glasses. This design allows for a slightly larger processor to support onboard inference without obstructing the visual field.

We used the ArduCam HM0360 VGA SPI camera due to its relatively high resolution, fast frame rate, and low power consumption (i.e., under 19.6 mW [53]). The low power consumption allows for an "always-on" operating mode for continuous visual detection and assessment of the walking environment. The camera frame rate of 60 fps can support environment-adaptive control of human-robot locomotion. The camera resolution (640 × 480) is larger than the input size of most deep learning models (e.g., MobileNetV2 has a default input of 224×224), while providing enough information to portray the environmental state.

We used the Raspberry Pi Pico W microcontroller for the onboard computational processing. This newly developed board offers increased memory and CPU power compared to smaller boards. Its enhanced processing power, large memory, small form factor, and wireless communication make it suitable for our design. The Pico contains Dual ARM 133 MHz processors, outperforming microcontrollers of comparable sizes, such as the Arduino Nano 33 BLE with a 64 MHz processing speed. This added processing power allows for increased speed and parallel processing to compute video stream and model inference. The Pico also has 64 kB SRAM and 2 MB QSPI flash memory, which is important as deep learning models must be stored on the embedded system to perform on-device inference. The Pico has a small form factor of 21 mm x 51.3 mm, which can more easily integrate into eyeglass frames. The microcontroller can also wirelessly communicate and interface with external robotic devices and computers via a single-band 2.4 GHz Wi-Fi connection or through Bluetooth 5.2.

We developed a deep learning model using a similar approach as our baseline model in Section 3.1. However, fine-tuning was required to convert the model from the chest-mounted domain to an eye-level domain. To do this, the baseline model was retrained using 7,250 images from the Meta Ego4D dataset [54] that we manually annotated, which contained walking environments that matched the StairNet classes (i.e., LG, LG-IS, IS, and IS-LG), with an input size of 96x96. We used the lightweight MobileNetV1 architecture to reduce the model size for embedded deployment compared to larger architectures like MobileNetV2. We performed hyperparameter optimization for batch size and learning rate with optimal values of 32 and 0.0001, respectively. The final model contained 219,300 parameters, was converted to a TensorFlow Lite model using int8 quantization and further reduced to a TensorFlow Micro model for deployment (Figures 5 and 6). We measured the onboard inference time as the loop of loading the most recent image captured and running the model inference directly on the microcontroller.

The average onboard inference speed was 1.47 seconds from reading the image to outputting the predicted label. Prior to domain fine-tuning, the model achieved a similar performance to our baseline model on the StairNet test set of 98.3% accuracy. Once fine-tuned with the Ego4D images from head-mounted cameras, the model could classify complex stair environments with 98.2% accuracy. To our knowledge, these AI-powered smart glasses are the first to integrate both sensing and computation for visual perception of human-robot walking environments.

4.  Discussion

In this study, we present a comprehensive overview of our StairNet initiative, which was created to support the development of new deep learning models for visual perception of stair environments for human-robot locomotion. The initiative places emphasis on efficient neural networks for onboard real-time inference on mobile and embedded devices. First, we outlined our development of the StairNet dataset with over 515,000 manually labeled images, followed by our development of different state-of-the-art deep learning models (e.g., 2D and 3D CNN, hybrid CNN and LSTM, and ViT networks) and training methods (e.g., supervised learning with and without temporal data, and semi-supervised learning with unlabeled images) using our new dataset. Our models consistently achieved high classification accuracy (i.e., up to 98.8%) with different designs, offering trade-offs between model size and performance. When deployed on mobile devices with GPU and NPU accelerators, our deep learning models achieved inference speeds up to 2.8 ms. When deployed on our CPU-powered smart glasses, which account for human-computer interaction, the inference speed was slower (i.e., 1.5 s). Overall, we showed that StairNet can serve as an effective platform to develop and study new visual perception systems for human-robot locomotion, with applications in environment-adaptive control of prosthetic legs, exoskeletons, and other mobility assistive technologies.

Our models offer several benefits over other stair recognition systems [6]–[11], [14]–[18],[25],[26]. Many previous studies have been limited to statistical pattern recognition and machine learning algorithms that require manual feature engineering. In contrast, our models use multilayer deep neural networks for automatic feature extraction, which has shown to be superior to handcrafted features [21]. Additionally, our models benefit from the high quantity and quality of the StairNet dataset, with over 515,000 manually annotated images, allowing for more generalizable systems. Previous research has mainly been limited to smaller datasets (see Table 1). These differences have important practical implications as the performance and use of machine learning models require large amounts of diverse data. The increased generalization potential of our models also eliminates the need for explicit requirements for the pose or angle of our camera, as observed in past studies that relied on meticulous rule-based thresholds for the dimensions of the user and environments [10]. Our system only offers general suggestions for the type of camera and mount location, which provides greater flexibility for future research.

We studied a wide variety of deep learning models and training methods (Table 2), each of which offer unique advantages and trade-offs. For example, the MoViNet 3D CNN using temporal data [30] achieved the highest classification accuracy on our StairNet test set compared to our baseline 2D CNN model, with a performance increase of 1.1%, demonstrating the benefit of temporal data for human-robot walking. However, the model contains a relatively large number of parameters (4.03 million) and numerical operations (2.5 GFLOPs), which could hinder deployment and real-time inference on mobile and embedded devices with limited computational resources; such models might be better suited for use cases with access to reliable cloud computing. For model efficiency, our MobileViT XS model trained using semi-supervised learning achieved the highest NetScore of 202.4 [41], demonstrating the benefit of using lightweight vision transformers to reduce model parameter count compared to standard convolutional neural networks. Additionally, our semi-supervised learning model showed the ability to use unlabeled data to improve training efficiency by reducing the number of annotated images required by 35% while maintaining the classification performance. The high efficiency of the MobileViT XS model makes it well-suited for our computer vision application.

We also studied different state-of-the-art edge devices through our development of a new mobile app [13] and smart glasses [49]. The mobile app uses a TFLite interpreter and on-device GPU and NPU accelerators to maximize inference. The app achieved fast inference speeds of 2.75 ms when running our baseline model. However, the mobile app used a chest-mounted smartphone with images that do not necessarily coincide with the user's visual field of view and are susceptible to obstructions such as the user's arms. This motivated our development of the smart glasses to improve human-computer interaction with an integrated design that takes into account the head orientation, thus having greater potential to infer the user's locomotor intent. However, limitations in the embedded system yielded slower inference speeds of 1.5 s, presenting a trade-off between human-centred design and performance. These slower inference speeds could affect the feasibility for real-time environment-adaptive control. Future work will thus focus on improving the onboard inference speed and further reducing the size of the hardware.

Despite these developments, our research has several limitations. First, to evaluate performance, we used the StairNet test set. Although test sets are common practice in deep learning [21], the true real-world performance and generalizability of our models was not analyzed in a deployed environment. Also, during the development of our temporal models, we identified a limitation of the training method used for our baseline and semi-supervised models as the train/validation/test splits were performed randomly between images. This caused data leakage between the different data subsets, with unintentionally higher classification performances for our baseline and semi-supervised models. Retesting revealed an updated baseline accuracy of 97.2% when using dataset splits with randomly sorted videos without neighboring frames in multiple data subsets. To address this, the performance evaluations were made based on the change in accuracy compared to our baseline model of the respective test set. For future development using our StairNet dataset, we suggest using these video-based training/validation/test splits.

It is important to also mention that state-of-the-art models and methods are continuously being developed. For example, during the course of our development of the temporal models, research on transformers [55] and multilayer perceptrons [56] showed the ability to eliminate the need to process each frame for the encoder and temporal blocks separately by adapting the models to take 3D sequence inputs by modifying the patch-embedding block, which can significantly improve the efficiency in processing and inference. For our semi-supervised learning research, many other algorithms besides FixMatch [42] could have been used to further reduce the number of labeled images required for stair recognition such as invariant semantic information clustering [57] and cross-level discrimination for unsupervised feature learning [58]. Our visual perception systems, especially the smart glasses, could also be extended to other applications such as providing sensory feedback to persons with visual impairments by leveraging some of the recent advances in large vision-language models [59].

Lastly, we want to emphasize that our visual perception systems as part of the StairNet initiative are meant to supplement, not replace, the existing intent recognition systems for human-robot walking that use mechanical, inertial, and/or EMG data. Our lab views computer vision as a means to improve the speed and accuracy of locomotion mode recognition by minimizing the search space of potential solutions based on the perceived walking environment. In future work, we plan to focus on sensor fusion of vision with EMG and/or inertial data to determine if and when vision can improve performance. In conclusion, we showed that StairNet can be an effective platform to develop and study new visual perception systems for human-robot locomotion with applications in control of prosthetic legs, exoskeletons, and other mobility assistive technologies.

## 5. Acknowledgements

We want to thank members of the Bionics Lab, a part of the Artificial Intelligence and Robotics in Rehabilitation Team at the KITE Research Institute, Toronto Rehabilitation Institute, for their help. This study was supported by the Walter and Maria Schroeder Institute for Brain Innovation and Recovery and the AGE-WELL Networks of Centres of Excellence program. We dedicate this study to the people of Ukraine in response to the 2022 Russian invasion and war.

Table 1. Summary of previous vision-based stair recognition systems for robotic leg prostheses and exoskeletons. The dataset size (i.e., the number of images) and test accuracy are only for the environment classes relating to level-ground walking and stair ascent. The systems are organized in terms of the test accuracy (%).

| Reference | Camera | Position | Dataset Size | Classifier | Computing Device | Test Accuracy |
|---|---|---|---|---|---|---|
| [11] | RGB | Waist | 7,284 | Convolutional neural network | NVIDIA Titan X | 99.6% |
| [10] | Depth | Chest | 170 | Heuristic thresholding and edge detector | Intel Core i5 | 98.8% |
| [9] | Depth | Leg | 8,455 | Support vector machine | Intel Core i7-2640M | 98.5% |
| StairNet | RGB | Chest | 515,452 | Convolutional neural network | Google Cloud TPU | 98.4% |
| [17] | Depth | Leg | 3,000 | Convolutional neural network | NVIDIA Quadro P400 | 96.8% |
| [8] | Depth | Leg | 109,699 | Cubic kernel support vector machine | Intel Core i7-2640M | 95.6% |
| [14] | RGB | Chest | 34,254 | Convolutional neural network | NVIDIA TITAN Xp | 94.9% |
| [15] | RGB | Head | 123,979 | Bayesian deep neural network | NVIDIA Jetson TX2 | 93.2% |
| [16] | RGB | Leg | 123,954 | Bayesian deep neural network | NVIDIA Jetson TX2 | 92.4% |
| [18], [25], [26] | RGB | Chest | 542,868 | Convolutional neural network | Google Cloud TPU | 70.8% |

Table 2. Summary of our stair recognition systems (StairNet). The models were evaluated based on image classification accuracy and efficiency (i.e., NetScore – higher is better). The systems are organized by model type. We tested supervised learning (SL) and semi-supervised learning (SSL) methods, and many-to-one (M1) and many-to-many (MM) temporal neural networks. The dataset sizes for our baseline and temporal neural networks were 515,452 labeled images, and 300,000 labeled images and 1.8 million unlabeled images for our semi-supervised learning networks.

| Type | Dataset size | Training approach | Architecture | Change in accuracy compared to baseline | NetScore | Model Parameters (millions) |
|---|---|---|---|---|---|---|
| Baseline Neural Network | 515,452 labeled | SL - Single frame | MobileNetV2 | 0% | 186.8 | 2.3 |
| Temporal Neural Networks* | | SL – M1 | MoViNet | +1.1% | 167.4 | 4.0 |
| | | SL – M1 | MobileNetV2 + LSTM | +0.1% | 132.1 | 6.1 |
| | | SL – M1 | MobileViT-XXS + LSTM | -0.2% | 155.0 | 3.4 |
| | | SL – MM | MobileNetV2 + LSTM | -26.5% | 120.1 | 6.0 |
| Semi-Supervised Neural Network | 300,000 labeled, 1.8M unlabeled | SSL – Fix Match | MobileViT-XS | +0.4% | 202.4 | 1.9 |
| | | SSL – Fix Match | MobileViT-XXS | -0.7% | 186.5 | 0.9 |
| | | SSL – Fix Match | MobileViT-S | -1.2% | 169.7 | 4.9 |

*Evaluated using the video-based train/validation/test split as described in Section 3.3

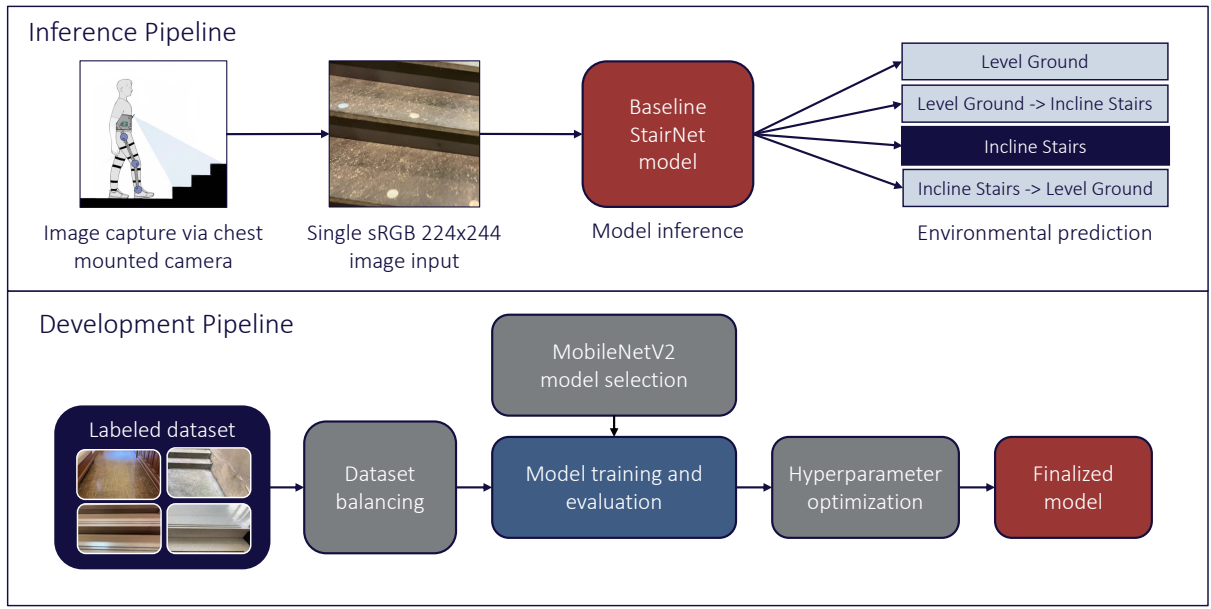

**Fig. 1** The inference and development pipelines for our baseline StairNet model [12] trained using supervised learning and single images for stair recognition. We developed this model as a reference and benchmark for our other deep learning models.

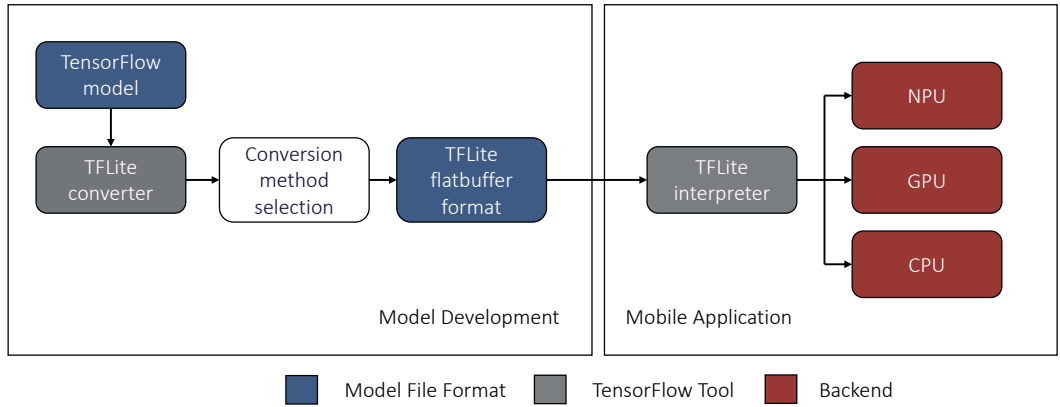

**Fig. 2** The model conversion and deployment pipeline for our mobile iOS application [13], which we developed to deploy and test our baseline model for on-device real-time inference.

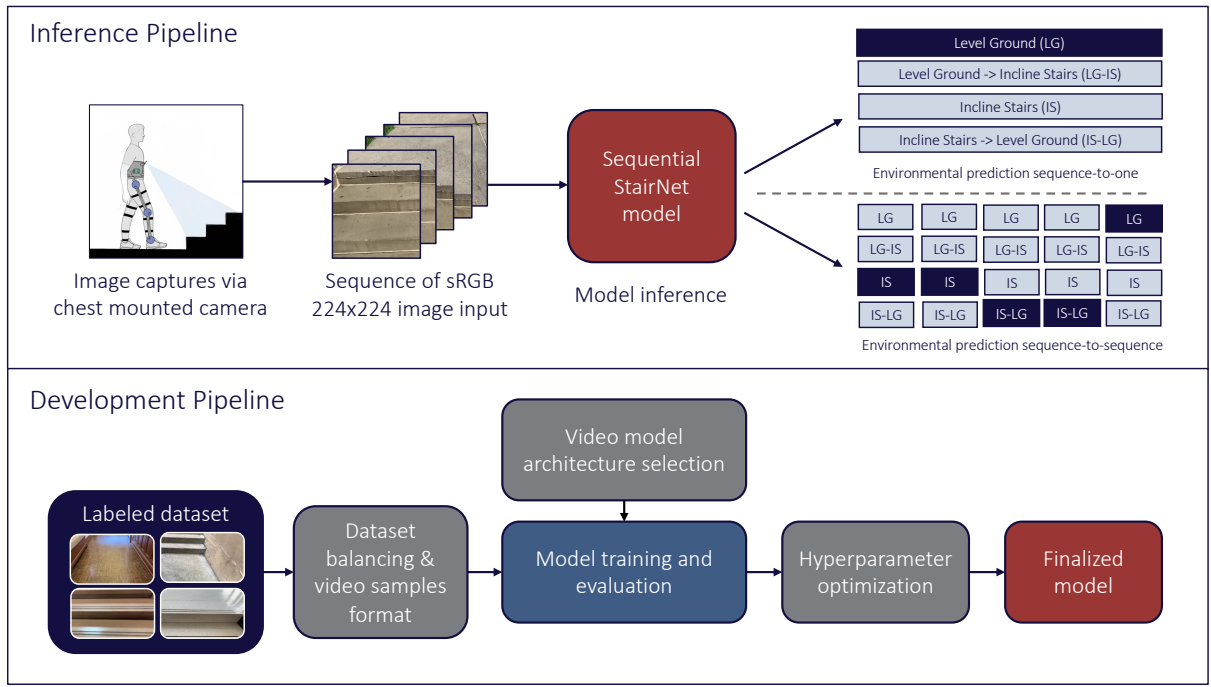

Fig. 3 The inference and development pipelines for our temporal StairNet models [30] trained using supervised learning and sequential images for stair recognition. Unlike our other models that used single image inputs, these temporal neural networks used sequential inputs.

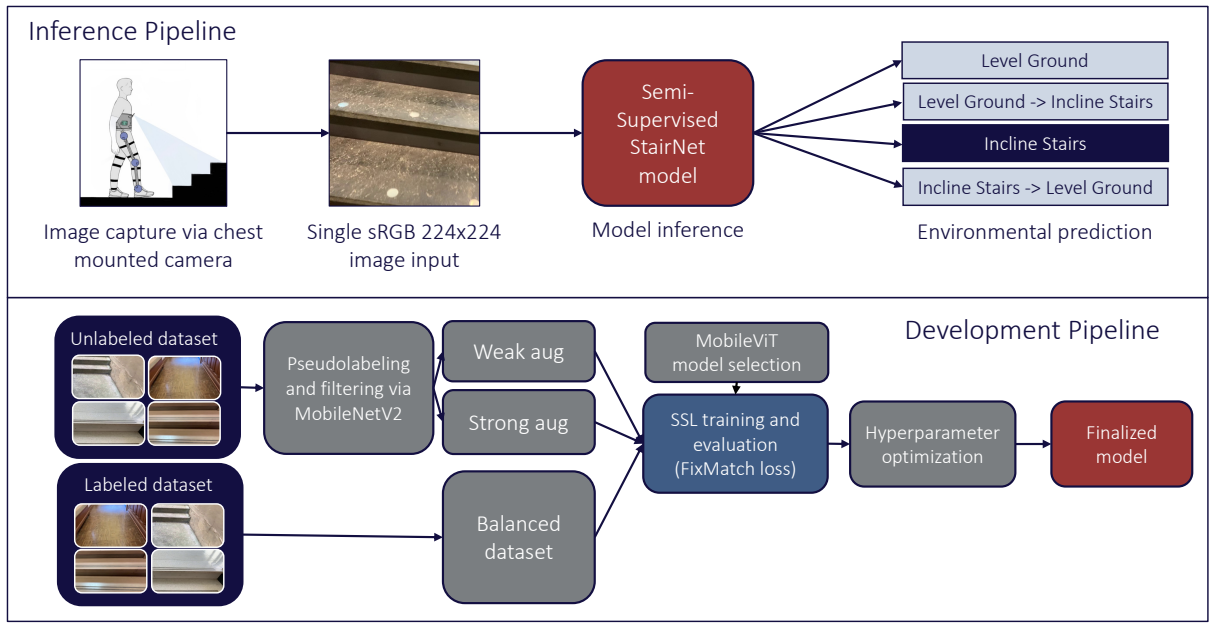

**Fig. 4** The inference and development pipelines for our semi-supervised learning StairNet model [41] trained using labeled and unlabeled images for stair recognition. Unlike our other models, this model used large amounts of unlabeled data to minimize annotation requirements while maintaining performance, thus improving training efficiency.

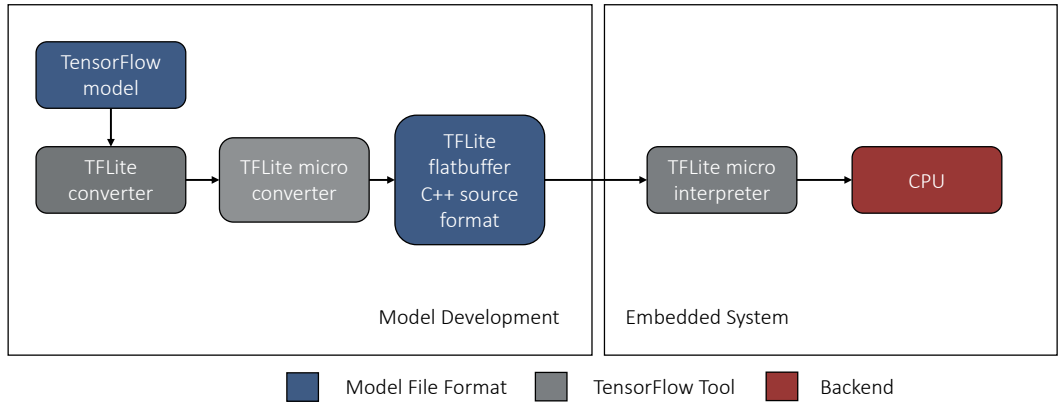

Fig. 5 The model conversion and deployment pipeline for our smart glasses [49], which we developed to deploy and test our model for real-time inference on an embedded device.

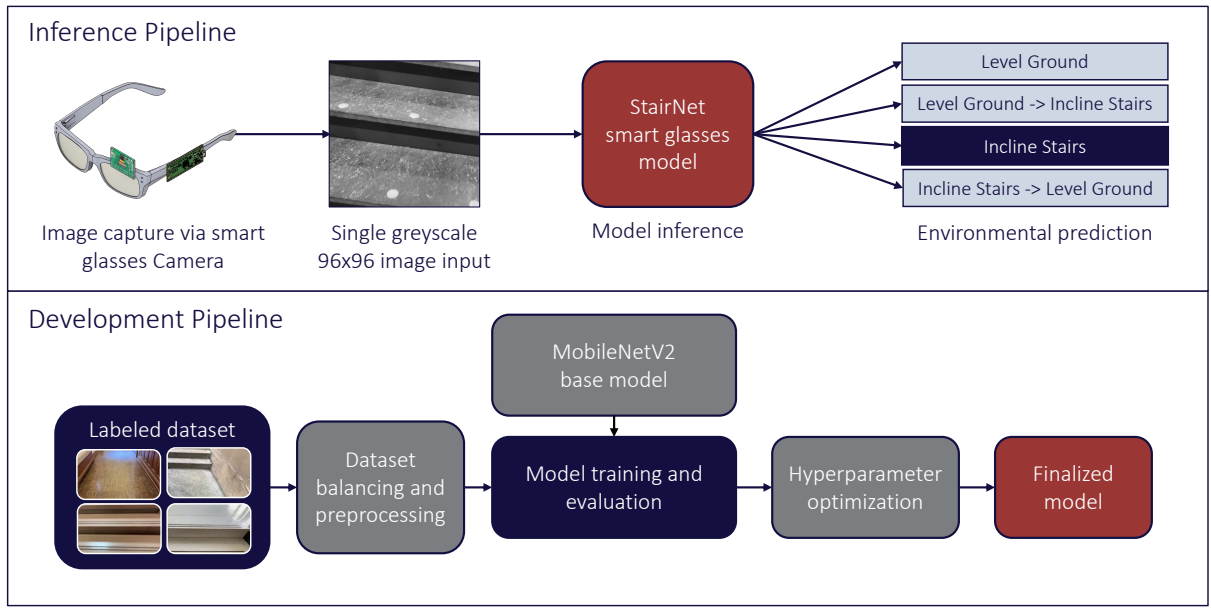

Fig. 6 The inference and development pipelines for our smart glasses StairNet model [49] trained using supervised learning and single images. Compared to our other models, the smart glasses performed stair recognition using a head-mounted camera and an embedded system.